\documentclass{article}

\newif\ifarxiv
\arxivtrue

\usepackage{natbib}
\usepackage[utf8]{inputenc} 
\usepackage[T1]{fontenc}    
\usepackage{url}            
\usepackage{booktabs}       
\usepackage{amsfonts}       
\usepackage{nicefrac}       
\usepackage{microtype}      
\usepackage{xcolor}         
\usepackage{fullpage}
\usepackage{kpfonts}
\usepackage{xcolor}
\definecolor{DarkGreen}{rgb}{0.1,0.5,0.1}
\definecolor{DarkRed}{rgb}{0.5,0.1,0.1}
\definecolor{DarkBlue}{rgb}{0.1,0.1,0.5}
\definecolor{Gray}{rgb}{0.2,0.2,0.2}
\usepackage[pdftex]{hyperref}

\usepackage{authblk}
\usepackage{wrapfig}
\usepackage[table,xcdraw]{xcolor}  
\usepackage[most]{tcolorbox}
\definecolor{PromptBg}{HTML}{C4E32B} 
\definecolor{RedColor}{rgb}{1, 0, 0}

\definecolor{PromptColor}{HTML}{6C8104}
\definecolor{RewardColor}{HTML}{4573B4}
\definecolor{PolicyColor}{HTML}{FEAB06} 
\definecolor{allcolor}{HTML}{F2EDFE}
\usepackage{multirow}

\hypersetup{
    unicode=false,          
    pdftoolbar=true,        
    pdfmenubar=true,        
    pdffitwindow=false,      
    pdfnewwindow=true,      
    colorlinks=true,       
    linkcolor=DarkBlue,          
    citecolor=DarkGreen,        
    filecolor=DarkRed,      
    urlcolor=DarkBlue,          
    pdftitle={How to Train Your Deep Research Agent? Prompt, Reward, and Policy Optimization in Search-R1},
    pdfauthor={Yinuo Xu, Shuo Lu, Jianjie Cheng, Meng Wang, Qianlong Xie, Xingxing Wang, Ran He, Jian Liang},
}

\usepackage{subcaption}
\usepackage{graphicx}

\title{How to Train Your Deep Research Agent? \\Prompt, Reward, and Policy Optimization in Search-R1}

\author{
  Yinuo Xu$^{\diamondsuit,\heartsuit}$ \thanks{Equal contribution.}\quad
  Shuo Lu$^{\diamondsuit}$ \protect\footnotemark[1]\quad
  Jianjie Cheng$^{\clubsuit}$ \quad
  Meng Wang$^{\clubsuit}$ \quad 
  Qianlong Xie$^{\clubsuit}$ \quad \\
  \vspace{-3mm}
  Xingxing Wang$^{\clubsuit}$ \quad 
  Ran He$^{\diamondsuit}$ \quad
  Jian Liang$^{\diamondsuit,\heartsuit}$ \thanks{Corresponding author.}\\
  \vspace{1mm}
  $^{\diamondsuit}$NLPR \& MAIS, CASIA \quad
  $^{\heartsuit}$School of AI, UCAS \quad
  $^{\clubsuit}$Meituan Inc. \\
  \vspace{1mm}
  \ttfamily\small
  \{YNfloxxxt, liangjian92\}@gmail.com
}

\date{}

\begin{document}

\maketitle

\begin{abstract}
Deep Research agents tackle knowledge-intensive tasks through multi-round retrieval and decision-oriented generation. 
While reinforcement learning (RL) has been shown to improve performance in this paradigm, its contributions remain underexplored.
To fully understand the role of RL, we conduct a systematic study along three decoupled dimensions: \textbf{prompt template}, \textbf{reward function}, and \textbf{policy optimization}.
Our study reveals that: 1) the Fast Thinking template yields greater stability and better performance than the Slow Thinking template used in prior work;
2) the F1-based reward underperforms the EM due to training collapse driven by answer avoidance; this can be mitigated by incorporating action-level penalties, ultimately surpassing EM;
3) REINFORCE outperforms PPO while requiring fewer search actions, whereas GRPO shows the poorest stability among policy optimization methods.
Building on these insights, we then introduce \textbf{Search-R1++}, a strong baseline that improves the performance of Search-R1 from 0.403 to 0.442 (Qwen2.5-7B) and 0.289 to 0.331 (Qwen2.5-3B).
We hope that our findings can pave the way for more principled and reliable RL training strategies in Deep Research systems.
\end{abstract}

\section{Introduction}
\label{sec:intro}

\begin{wrapfigure}{r}{0.38\textwidth}
    \centering
    \vspace{-15mm}
    \includegraphics[width=\linewidth]{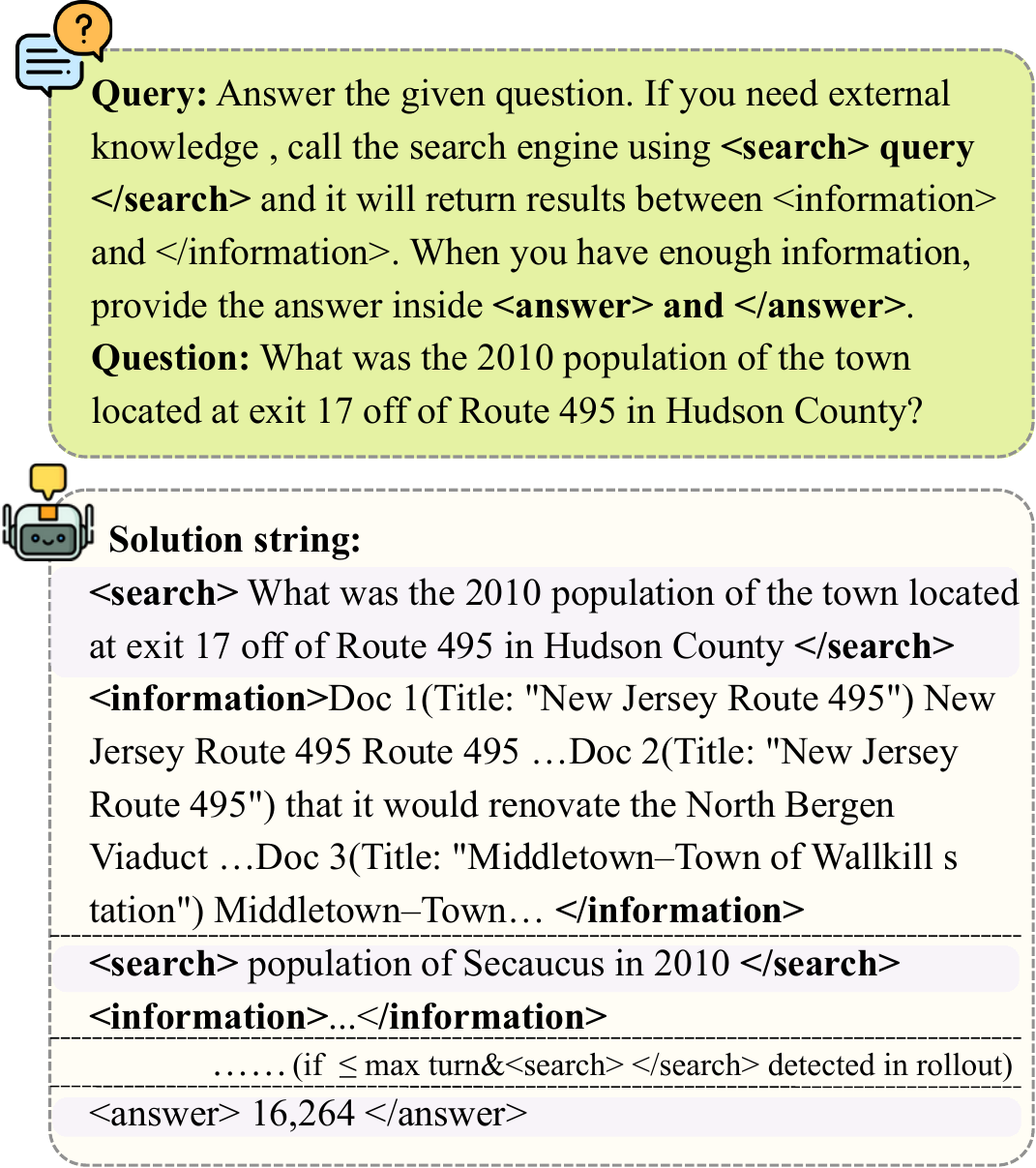}
    \caption{An example of the Search-R1 generation pipeline.}
    \vspace{-10mm}
    \label{fig:demo}
\end{wrapfigure}

\begin{figure*}[t]
    \centering
    \includegraphics[width=0.85\textwidth]{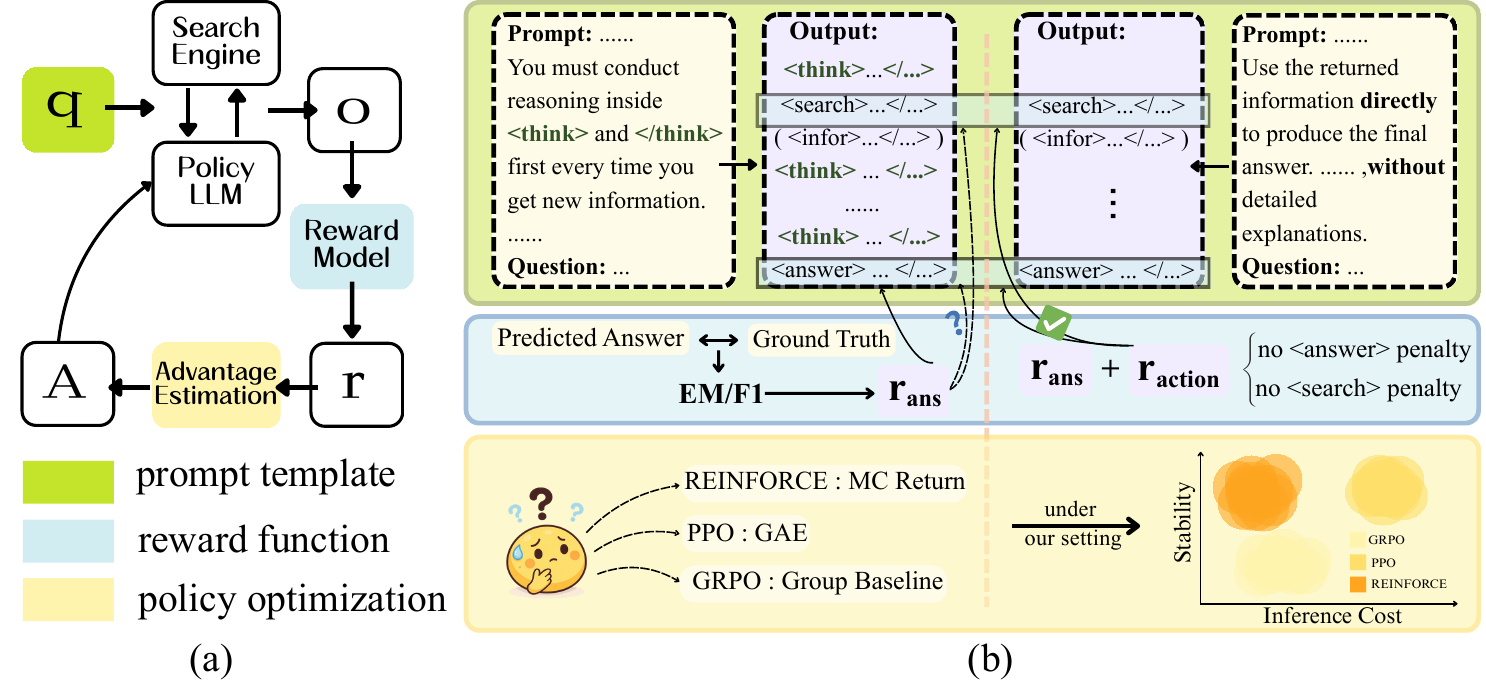}
    \caption{(\textbf{a}) demonstrates Deep Research's RL training pipeline; (\textbf{b}) shows an overview of the three key aspects explored in our work: \textbf{prompt template}, \textbf{reward function}, and \textbf{policy optimization}.}
    \label{fig:overview}
\end{figure*}

Deep Research agents have emerged as a dominant paradigm for solving complex knowledge-intensive tasks~\citep{xu2025comprehensive,huang2025deepresearchagentssystematic}. As shown in Figure~\ref{fig:demo}, they operate through multi-round retrieval, evidence aggregation, and decision-oriented generation~\citep{jin2025search,lewis2020retrieval}.
Reinforcement learning (RL) naturally fits this setting, as it directly optimizes long-horizon interactive behaviors under sparse feedback~\citep{silver2016mastering}, thereby reducing reliance on dense expert search trajectories that SFT depends on~\citep{li2025reinforcementlearningfoundationsdeep}.

Despite its growing dominance, RL training recipes in Deep Research remain fragmented, making it difficult to identify which configurations really drive performance gains.
To systematically assess the role of RL, we adopt a unified framework that focuses on three primary dimensions: prediction accuracy, training stability, and inference cost. Within this unified analytical framework as shown in Figure~\ref{fig:overview}, we then dissect how these dimensions are shaped by three decoupled components of the overall training pipeline: \textbf{prompt template}, \textbf{reward function}, and \textbf{policy optimization}.

First, we investigate the role of prompt design in optimizing Deep Research agents, a component shown to be significant for RL training in related systems~\citep{zhou2025gui,deng2022rlprompt}. 
Experiments on existing Search‑R1–style systems reveal that longer reasoning trajectories or richer external information do not necessarily lead to more reliable performance.
Motivated by this finding, we introduce the Fast Thinking template, which encourages the policy to output search and answer decisions directly during training. 
When compared to the widely adopted Slow Thinking template, the Fast Thinking template produces more stable training and achieves better performance.

Second, while current Deep Research systems have shifted from Exact Match (EM) rewards~\citep{cheng2025agentr1trainingpowerfulllm,jin2025search} to near-standard reliance on F1 scores~\citep{sun2025zerosearch,zheng-etal-2025-deepresearcher}, our experiments show that F1-based training exhibits inferior stability and performance. We further identify that training collapse mainly stems from answer avoidance, where the policy learns to withhold final answers rather than produce incorrect ones, revealing that sole F1 supervision fails to provide sufficient constraints on intermediate actions.  
By augmenting the reward with lightweight action-level penalties, we  mitigate this collapse, improving stability and enabling F1-based training to surpass the EM-based counterpart.

Third, we systematically compare the three policy optimization algorithms most prevalent in Deep Research: PPO, GRPO, and REINFORCE~\citep{sun2025zerosearch,jin2025search}. 
Our experiments reveal that GRPO exhibits the poorest training stability. 
Notably, despite being an older and simpler algorithm, REINFORCE achieves superior final performance and greater efficiency than both PPO and GRPO, converging with better output accuracy fewer search actions.

Building on these insights, we introduce \textbf{Search-R1++}, which utilizes the Fast Thinking template and is trained via REINFORCE with F1 rewards augmented by action-level penalties.
On standard Deep Research benchmarks, \textbf{Search-R1++}  substantially outperforms the Search-R1 baseline, raising the average accuracy from 0.403 to 0.442 (Qwen2.5-7B) and 0.289 to 0.331 (Qwen2.5-3B).

To summarize, we present the first comprehensive study investigating the role of RL in Deep Research. 
Our key findings demonstrate: 1) the Fast Thinking template yields greater stability and better performance than the widely adopted Slow Thinking template; 
2) augmenting F1 rewards with action-level penalties prevents training collapse and enables them to outperform EM; 
3) while GRPO shows the poorest stability, the classic REINFORCE algorithm performs best with greater efficiency.
Building on these insights, we establish a strong baseline: \textbf{Search-R1++}.
We hope this work inspires more principled and reliable RL training strategies for Deep Research, shifting the optimization focus from simply integrating richer tools or adopting newer algorithms toward more deliberate component design.

\section{Deep Research}
\label{sec:bg}

\noindent \textbf{Paradigms.} Deep Research agents conduct multi-round retrieval, evidence aggregation, and decision-oriented generation to automate external information seeking in knowledge-intensive settings such as open-domain QA and long-document summarization~\citep{openaideepresearch,nakano2021webgpt,lewis2020retrieval}. Prompt-based agents use structured prompts to explicitly guide agents to perform iterative reasoning and retrieval~\citep{asai2024self,trivedi2023interleaving,yao2022react}. SFT-based methods imitate human or rule-generated search trajectories to learn when to retrieve, what to query, and how to integrate evidence~\citep{schick2023toolformer,wang2025chainofretrievalaugmentedgeneration,yu2024rankrag}. More recently, RL-based methods treat Deep Research as a sequential decision problem and directly optimize search and answering strategies over long-horizon interactions~\citep{zheng-etal-2025-deepresearcher,song2025r1,sun2025zerosearch,chen2025learning,lu2025deepresearch}. In this setting, the agent repeats “think–search–rethink” steps under sparse feedback, making this long-horizon interactive process naturally aligned with RL~\citep{kumartraining,nachum2016improving,ladosz2022exploration}. However, despite the growing popularity of RL for Deep Research, existing studies often adopt custom implementations and varied configurations. This results in highly fragmented training recipes that obscure which factors truly drive stronger performance. 

\noindent \textbf{Framework.} Search-R1~\citep{jin2025search} serves as a representative baseline for RL-driven Deep Research. In this study, we strictly replicate the Search-R1 architecture, including its generation pipeline, dataset, and retriever, to isolate the impact of different RL training settings within this controlled environment. To facilitate a systematic analysis, we decompose the training pipeline into three decoupled aspects: \textbf{prompt template}, \textbf{reward function}, and \textbf{policy optimization}, and comprehensively evaluate the agent's capabilities across metrics of prediction accuracy, training stability, and inference cost~\citep{ouyang2022training,nakano2021webgpt}. Detailed experimental findings and analyses are presented in the subsequent sections.
\section{Prompt Template}
\label{sec:prompt}

\subsection{Setup}
\label{sec:data}

\begin{wrapfigure}{r}{0.4\textwidth}
    \centering
    \vspace{-12mm}
    \includegraphics[width=0.9\linewidth]{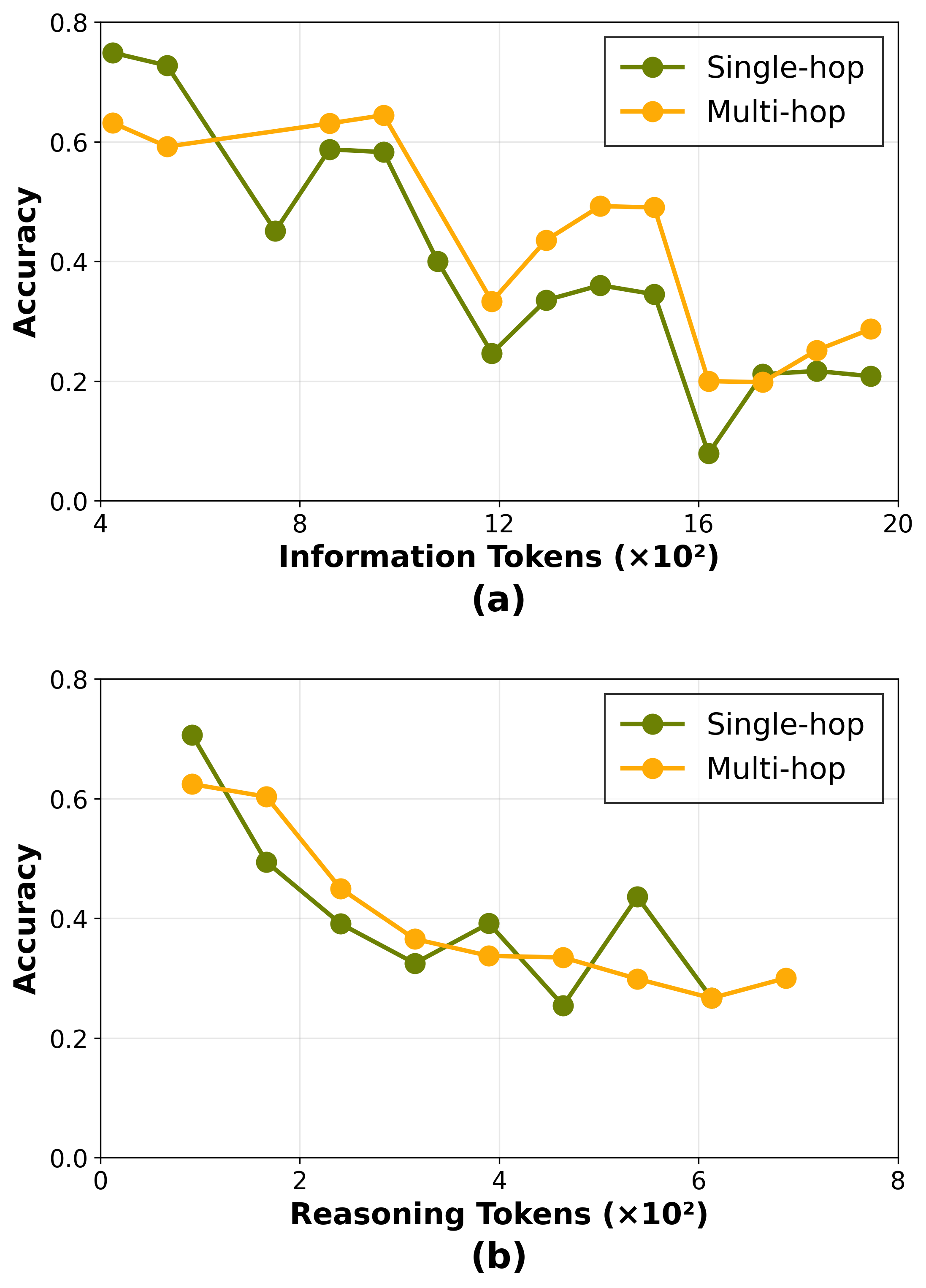}
    \caption{(\textbf{a}) Accuracy under varying information tokens; (\textbf{b}) Accuracy under varying reasoning tokens.}
    \vspace{-15mm}
    \label{fig:prompt-1}
\end{wrapfigure}

\textbf{Experimental Setup}. In this section, following the standard Search-R1~\citep{jin2025search} framework, we perform experiments using Qwen2.5-3B and Qwen2.5-7B~\citep{qwen2025qwen25technicalreport}, both trained via PPO with EM reward. Retrieval uses E5~\citep{wang2022text} on 2018 Wikipedia~\citep{karpukhin2020dense}, retrieving top-3 relevant passages.  Unless stated otherwise, we perform the analysis using Qwen2.5-7B. More detailed experimental settings and additional results are provided in the Appendices~\ref{sec:appendix_3b} and~\ref{sec:detail setup}.

\noindent{\textbf{Datasets and Evaluation}}. Our study utilizes seven established benchmarks categorized into two groups: (1) \textbf{General QA}, comprising NQ \citep{kwiatkowski2019natural}, TriviaQA \citep{joshi2017triviaqa}, and PopQA \citep{mallen2023not}; and (2) \textbf{Multi-Hop QA}, including HotpotQA \citep{yang2018hotpotqa}, 2WikiMultiHopQA \citep{ho2020constructing}, Musique \citep{trivedi2022musique}, and Bamboogle \citep{press2023measuring}. Specifically, we merge the standard training splits of NQ and HotpotQA for training, and conduct evaluations on the respective test or validation sets of all seven datasets. Exact Match (EM) score is used as the primary metric to evaluate final answer accuracy.

\subsection{The Less Thinking, the Better Performance}

\begin{figure*}[t]
    \centering
    \includegraphics[width=0.99\textwidth]{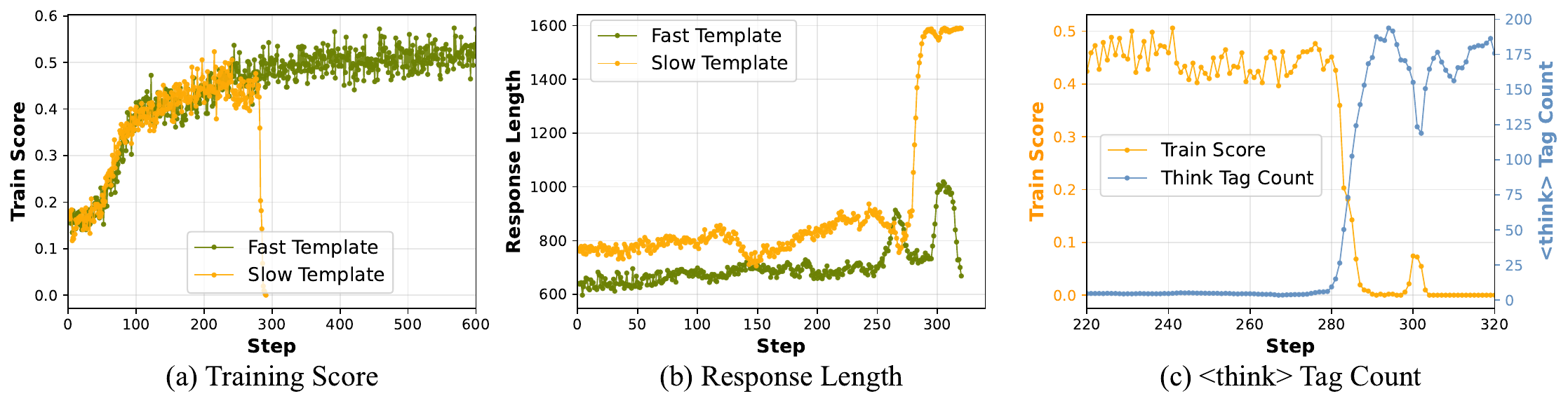}
    \caption{(\textbf{a}) compares the training score under Fast and Slow Thinking templates; (\textbf{b}) shows the average response length evolution over training steps; (\textbf{c}) illustrates the surge in \texttt{<think>} tags coinciding with the performance collapse.}
    \label{fig:prompt-2}
\end{figure*}

Prompt design is a critical factor in optimizing RL-based systems, as instruction strategies significantly influence training efficiency and agent behavior. In the context of reinforcement learning, the structural constraints of a prompt define the exploration boundaries of the agent and directly impact the density of the feedback signal during the policy optimization process~\citep{ouyang2022training}. While extended reasoning chains are traditionally seen as beneficial for System-2 tasks such as mathematics and program synthesis~\citep{guo2025deepseek,wei2022chain,lewkowycz2022solving,chen2022program}, recent studies have reported that introducing explicit intermediate reasoning can degrade performance in tasks such as complex reasoning and image grounding tasks~\citep{li2025think,zhang2025does}. In the context of Deep Research agents, systems like Search-R1 commonly introduce explicit reasoning by using think-before-search/action prompts, with dedicated \texttt{<think>} tags guiding model reasoning, as illustrated in Figure~\ref{fig:overview}(b). Although these reasoning-augmented templates achieve strong empirical performance, it remains unclear whether these gains truly arise from the reasoning process itself. 

Before diving into the analysis, we first formalize the model’s input and output for consistency across experiments. The input consists of a question $q$ and an instruction prompt $t_{\mathrm{ins}}$. A full rollout includes explicit reasoning process $t_{\mathrm{think}}$ enclosed in \texttt{<think>...</think>}, search queries $t_{\mathrm{query}}$, retrieved information $t_{\mathrm{info}}$, and a final answer $t_{\mathrm{ans}}$. We define the total number of all information tokens $t_{\mathrm{info}}$ as information tokens, and the total number of reasoning tokens $t_{\mathrm{think}}$ as reasoning tokens.

\textbf{Longer explicit reasoning leads to worse task performance.} Based on our formalized metrics, we perform a statistical analysis on Search-R1 (Qwen2.5-7B) across a suite of general question answering benchmarks covering both single-hop and multi-hop settings. As shown in Figure~\ref{fig:prompt-1}, contrary to the intuition that more thought improves outcomes, increases in explicit reasoning and retrieved information consistently correlate with lower accuracy in both settings. While acknowledging the potential influence of confounding factors, this pattern demonstrates a clear negative correlation, suggesting that more complex reasoning does not necessarily translate into better task performance.

\subsection{Fast vs. Slow Thinking Templates}
Building on the above observations, we evaluate the Fast Thinking template and the Slow Thinking template under the same training configuration as illustrated in section~\ref{sec:data} and appendix~\ref{sec:detail setup}. We provide detailed rollout cases for different prompts in the Appendix~\ref{sec:case}, exemplified by Qwen2.5-7B.

\begin{tcolorbox}[
    colback=PromptBg!20,   
    colframe=black,        
    arc=10pt,              
    boxrule=0.8pt,         
    left=4pt,right=4pt,top=4pt,bottom=4pt,
    breakable,             
    fontupper=\small,      
    before upper={\setlength{\parskip}{1pt}} 
]
\noindent
\textbf{Template 1 (Fast Thinking Template). }Answer the given question. If you need external knowledge to answer the question, call the search engine using \texttt{<search>} query \texttt{</search>} and it will return the top searched results between \texttt{<information>} and \texttt{</information>}. You can search as many times as you want. Use the returned information directly to produce the final answer. When you have enough information, provide the answer inside \texttt{<answer>} and \texttt{</answer>}, without detailed explanations. For example, \texttt{<answer>xxx</answer>}. Question: \textcolor{red}{question}.

\vspace{0.5em}

\noindent
\textbf{Template 2 (Slow Thinking Template). }
Answer the given question. You must conduct reasoning inside \texttt{<think>} and \texttt{</think>} first every time you get new information. After reasoning, if you find you lack some knowledge, you can call a search engine by \texttt{<search>} query \texttt{</search>} and it will return the top searched results between \texttt{<information>} and \texttt{</information>}. You can search as many times as you want. If you find no further external knowledge needed, you can directly provide the answer inside \texttt{<answer>} and \texttt{</answer>}, without detailed illustrations. For example, \texttt{<answer>xxx</answer>}. Question: \textcolor{red}{question}.
\end{tcolorbox}

\begin{table}[!ht]
\centering
\caption{Performance comparison of Slow vs.\ Fast Thinking templates across different model sizes. The Slow Thinking template corresponds to our reproduction of Search-R1~\citep{jin2025search}.}
\label{tab:prompt}
\small
\setlength{\tabcolsep}{5pt}
\renewcommand{\arraystretch}{1.1} 
\resizebox{\textwidth}{!}{
\begin{tabular}{lcccccccc}
\toprule
\multirow{2}{*}{\textbf{Method}} 
& \multicolumn{3}{c}{\textbf{Single-Hop QA}} 
& \multicolumn{4}{c}{\textbf{Multi-Hop QA}} 
& \multirow{2}{*}{\textbf{Avg.}} \\
\cmidrule(lr){2-4} \cmidrule(lr){5-8}
& \textbf{NQ} & \textbf{TriviaQA} & \textbf{PopQA} 
& \textbf{HotpotQA} & \textbf{2Wiki} & \textbf{Musique} & \textbf{Bamboogle} & \\
\midrule
\rowcolor{gray!15}\multicolumn{9}{c}{Qwen2.5-7B} \\
Slow Thinking template (Search-R1)
& 0.451 & 0.620 & 0.434 & 0.361 & \textbf{0.386} & \textbf{0.163} & 0.406 & 0.403 \\
Fast Thinking template (Ours)
& \textbf{0.463} & \textbf{0.640} &\textbf{0.458} & \textbf{0.427} & 0.360 & 0.156 & \textbf{0.453} & \textbf{0.422} \\
\midrule
\rowcolor{gray!15}\multicolumn{9}{c}{Qwen2.5-3B} \\
Slow Thinking template (Search-R1)
& \textbf{0.396} & 0.570 & 0.381 & 0.263 & 0.254 & \textbf{0.048} & 0.109 & 0.289 \\
Fast Thinking template (Ours)
& 0.390 & \textbf{0.576} & \textbf{0.393} & \textbf{0.282} & \textbf{0.272} & 0.041 & \textbf{0.125} & \textbf{0.297} \\
\bottomrule
\end{tabular}
}
\vspace{0.1in}
\end{table}

\begin{table*}[!ht]
\centering
\caption{Performance comparison of different reward functions based on Qwen2.5-7B. F1+ denotes F1 reward augmented with penalties.}
\label{tab:reward_qwen7b}
\small
\setlength{\tabcolsep}{4.5pt}
\renewcommand{\arraystretch}{1.1}
\begin{tabular}{lcccccccc}
\hline
\multicolumn{1}{c}{\multirow{2}{*}[-0.8ex]{\begin{tabular}{@{}c@{}}\textbf{Method}\\[-3pt] \textbf{(Qwen2.5-7B)}\end{tabular}}}
& \multicolumn{3}{c}{\rule{0pt}{10pt}\textbf{Single-Hop QA}}
& \multicolumn{4}{c}{\textbf{Multi-Hop QA}}
& \multirow{2}{*}{\textbf{Avg.}} \\
\cmidrule(lr){2-4} \cmidrule(lr){5-8}
& \textbf{NQ} & \textbf{TriviaQA} & \textbf{PopQA}
& \textbf{HotpotQA} & \textbf{2Wiki} & \textbf{Musique} & \textbf{Bamboogle} & \\
\hline
\rowcolor{gray!10} \multicolumn{9}{c}{\rule{0pt}{10pt}\text{Evaluation Metric: Exact Match (EM)}} \\
F1 
& 0.404 & 0.634 & 0.406 & 0.382 & 0.392 & 0.148 & 0.375 & 0.391 \\
EM 
& 0.463 & 0.640 & \textbf{0.458} & 0.427 & 0.360 & 0.156 & \textbf{0.453} & 0.422 \\
F1+ 
& \textbf{0.483} & \textbf{0.659} & 0.441 & \textbf{0.434} & \textbf{0.401} & \textbf{0.159} & 0.422 & \textbf{0.429} \\
\hline
\rowcolor{gray!15}\multicolumn{9}{c}{\rule{0pt}{10pt}Evaluation Metric: F1 Score} \\
F1
& 0.506 & 0.709 & 0.484 & 0.495 & 0.416 & 0.245 & 0.445 & 0.471 \\
EM 
& 0.570 & 0.711 & \textbf{0.512} & 0.530 & 0.422 & 0.236 & 0.489 & 0.496 \\
F1+ 
& \textbf{0.588} & \textbf{0.736} & 0.508 & \textbf{0.554} & \textbf{0.485} & \textbf{0.260} & \textbf{0.545} & \textbf{0.525} \\
\hline
\end{tabular}
\end{table*}
\begin{table*}[!ht]
\centering
\caption{Performance comparison of different reward functions based on Qwen2.5-3B. F1+ denotes F1 reward augmented with penalties.}
\label{tab:reward_qwen3b}
\small
\setlength{\tabcolsep}{4.5pt}
\renewcommand{\arraystretch}{1.1}
\begin{tabular}{lcccccccc}
\hline
\multicolumn{1}{c}{\multirow{2}{*}[-0.8ex]{\begin{tabular}{@{}c@{}}\textbf{Method}\\[-3pt] \textbf{(Qwen2.5-3B)}\end{tabular}}}
& \multicolumn{3}{c}{\rule{0pt}{10pt}\textbf{Single-Hop QA}}
& \multicolumn{4}{c}{\textbf{Multi-Hop QA}}
& \multirow{2}{*}{\textbf{Avg.}} \\
\cmidrule(lr){2-4} \cmidrule(lr){5-8}
& \textbf{NQ} & \textbf{TriviaQA} & \textbf{PopQA}
& \textbf{HotpotQA} & \textbf{2Wiki} & \textbf{Musique} & \textbf{Bamboogle} & \\
\hline
\rowcolor{gray!15}\multicolumn{9}{c}{\rule{0pt}{10pt}Evaluation Metric: Exact Match (EM)} \\
F1 
& 0.402 & 0.565 & 0.380 & 0.261 & 0.245 & 0.038 & 0.125 & 0.288 \\
EM 
& 0.390 & 0.576 & 0.393 & 0.282 & 0.272 & 0.041 & 0.125 & 0.297 \\
F1+ 
& \textbf{0.421} & \textbf{0.600} & \textbf{0.431} & \textbf{0.292} & \textbf{0.290} & \textbf{0.060} & \textbf{0.156} & \textbf{0.321} \\
\hline
\rowcolor{gray!15}\multicolumn{9}{c}{\rule{0pt}{10pt}Evaluation Metric: F1 Score} \\
F1 
& 0.482 & 0.638 & 0.422 & 0.340 & 0.296 & 0.084 & 0.188 & 0.350 \\
EM
& 0.460 & 0.637 & 0.424 & 0.345 & 0.308 & 0.089 & 0.200 & 0.352 \\
F1+
& \textbf{0.497} & \textbf{0.662} & \textbf{0.488} & \textbf{0.383} & \textbf{0.345} & \textbf{0.123} & \textbf{0.302} & \textbf{0.400} \\
\hline
\end{tabular}
\end{table*}

\textbf{Training Stability.} Fast Thinking template exhibits more stable training dynamics, whereas Slow Thinking template is more likely to lead to training collapse, as shown in Figure~\ref{fig:prompt-2}(a). To understand this stability gap, we examine how the policy evolves during training. Before collapse, runs with the Slow Thinking template consistently produce longer responses than those with the Fast Thinking template, with a marked jump around the collapse point, as shown in Figure~\ref{fig:prompt-2}(b). In the same phase, we also observe a surge in the number of \texttt{<think>} tags, as shown in Figure~\ref{fig:prompt-2}(c). Inspecting specific rollout trajectories immediately prior to collapse, we observe that the model begins to generate multiple \texttt{<think>} segments before a single action decision and  degenerates into repeatedly outputting empty \texttt{<think></think>} blocks, which crowd out normal reasoning and decision making. We provide a detailed analysis of a representative pre-collapse case in Appendix~\ref{sec:case}.

To further analyze what drives this behavior, as detailed in Appendix~\ref{sec:person}, we compute the Pearson correlation between the number of \texttt{<think>} tags and the immediate reward within the 10 steps preceding the collapse step. The collapsing run exhibits a moderate positive correlation of 0.4310, compared to a near-zero correlation of -0.0465 under stable training. This observation suggests that during the collapse phase, the policy identifies a shortcut where an increased frequency of \texttt{<think>} tags is associated with higher rewards. Under the sparse reward structures typical of PPO, this can potentially lead to a faulty unintended credit assignment bias, where the agent learns to "stack" tags to maximize episodic returns. This self-reinforcing loop tendency eventually triggers the uncontrolled growth of explicit reasoning and leads to the observed final training collapse.

\textbf{Performance.} In terms of final performance, simply switching to the Fast Thinking template yields higher accuracy than the Slow Thinking template, improving overall correctness from 0.403 to 0.422 on Qwen2.5-7B and from 0.289 to 0.297 on Qwen2.5-3B, as shown in Table~\ref{tab:prompt}. 

\textbf{Slow Thinking encourages reasoning expansion for rewards, undermining training stability and final quality.} The explicit introduction of \texttt{<think>} tags encourages the model to optimize by extending unnecessary reasoning length rather than improving key decisions like search and answer. As training progresses and policy learning encounters its training bottleneck, the policy tends to accumulate excessive \texttt{<think>} tags, causing uncontrolled reasoning growth and ultimately leading to severe instability and reduced task performance. In contrast, the Fast Thinking Template effectively limits the unchecked expansion of reasoning, focusing policy updates more strictly on key decisions, thus achieving more balanced outcomes.

\section{Reward Function}
\label{sec:reward}

\subsection{Setup}
In this section, we conduct experiments with Qwen2.5-3B and Qwen2.5-7B trained via PPO with the Fast Thinking template, with detailed training settings in Appendix~\ref{sec:detail setup}. We mainly focus our analysis on Qwen2.5-7B, with more results for Qwen2.5-3B provided in Appendix~\ref{sec:appendix_3b}.

\subsection{Is F1 really better than EM?}
Given that existing approaches mainly rely solely on outcome-based rewards (e.g., EM or F1)~\citep{sun2025zerosearch,zheng-etal-2025-deepresearcher}, we first analyze their impact on training dynamics, revealing two observations.

First, training with the F1 score exhibits significantly higher instability compared to the Exact Match (EM) reward, and tends to produce longer answer sequences. As detailed in the statistics below, we report the Mean Length and 90th Percentile from the initial stable phase (first 250 steps), which characterize the average and the length threshold covering 90\% of the generated answer sequences:
\begin{center}
\begin{tabular}{lcc}
\toprule
\textbf{Reward} & \textbf{Mean Length} & \textbf{90th Percentile} \\
\midrule
F1 (train) & 2.85 & 4.0 \\
EM (train) & 2.42 & 3.0 \\
\bottomrule
\end{tabular}
\end{center}

Second, and more critically, this length bias negatively impacts final performance. Contrary to the expectation that optimizing F1 should yield better F1 scores, we observe that the model trained with the stricter EM reward consistently outperforms the F1-trained model on both EM and F1 validation benchmarks, as shown in Table~\ref{tab:reward_qwen7b} and Table~\ref{tab:reward_qwen3b}. 

\subsection{Why Does Training Collapse?}
\begin{figure}[b]
    \centering
    \begin{minipage}{0.48\textwidth}
        \centering
        \includegraphics[width=\linewidth]{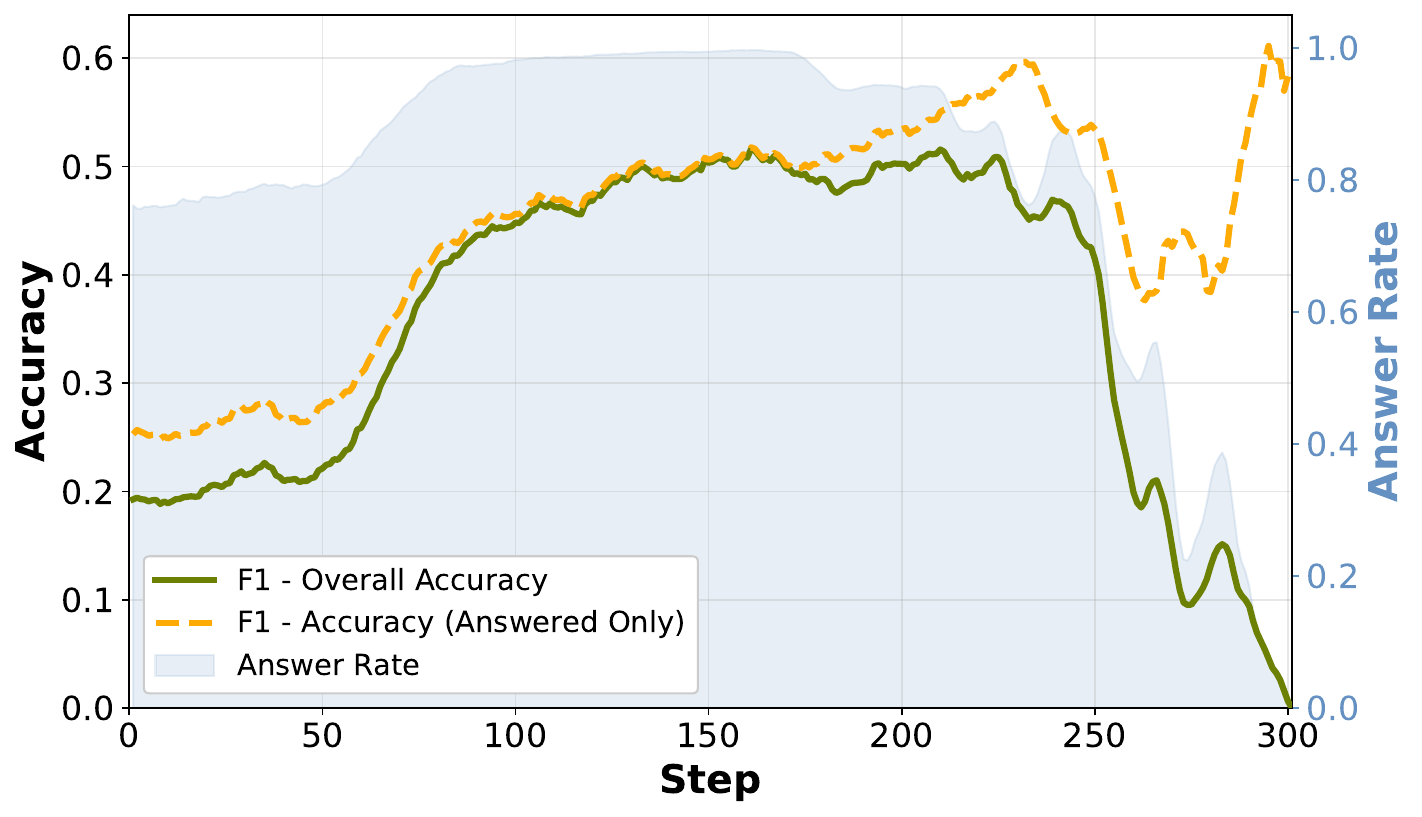}
        \caption{Overall accuracy, answered-only accuracy, and answer rate (shaded area) under F1 reward.}
        \label{fig:reward-1}
    \end{minipage}
    \hfill
    \begin{minipage}{0.48\textwidth}
        \centering
        \includegraphics[width=\linewidth]{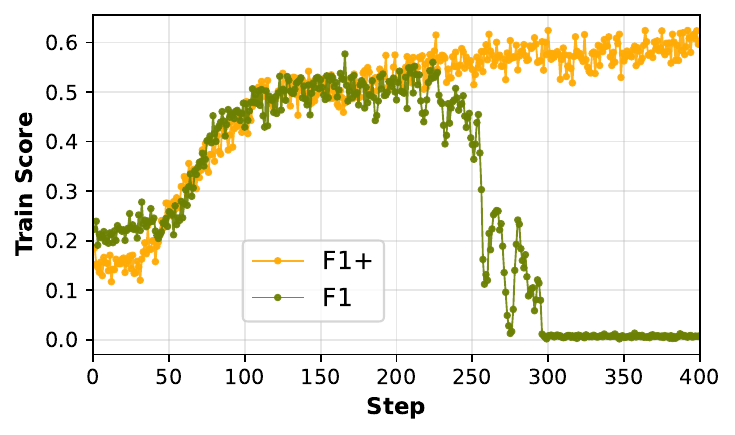}
        \caption{Training score under F1 and F1+ reward. F1+ denotes F1 augmented with action supervision.}
        \label{fig:reward-2}
    \end{minipage}
\end{figure}

To investigate the cause of the instability in F1-based training, we analyze the policy's behavior during failure phases and identify a consistent degeneration pattern: sharp drops in the overall score coincide with a significant decline in the answer rate, whereas the accuracy of the answered samples remains relatively stable, as shown in Figure~\ref{fig:reward-1}. This indicates that the dominant failure mode is not incorrect answering, but rather answer avoidance. Specifically, sole outcome-based supervision lacks sufficient constraints on the decision-making process. Since missing an answer receives the same zero reward as an incorrect one, the policy tends to collapse into a simpler state of generating no answer, effectively avoiding the effort of complex reasoning.

\subsection{Revitalizing F1 through Action Supervision}

Based on the insight that the agent tends to bypass critical decisions to minimize risk, we augment the outcome reward with minimal action-level penalties to discourage omitting search or answer steps. We denote this augmented reward as \textbf{F1+}, defined as:
\begin{equation}
\label{eq:f1_plus_reward}
R_{\text{F1+}} = R_{\text{F1}} - \alpha \cdot \mathbb{I}[a_s = 0] - \beta \cdot \mathbb{I}[a_a = 0],
\end{equation}
where $R_{\text{F1}}$ is the standard F1-based outcome reward, $a_s$ is the number of search actions executed in the step, $a_a$ is the number of answers produced, $\mathbb{I}[\cdot]$ is the indicator function, and we set $\alpha = 0.1, \beta = 0.1$ here.

While explicitly forcing actions carries a theoretical risk of reward hacking~\citep{fu2025reward}, our empirical results show a positive effect. These constraints stabilize training dynamics by eliminating the answer refusal failure mode, as shown in Figure~\ref{fig:reward-2}. Most importantly, as shown in Table~\ref{tab:reward_qwen7b} and Table~\ref{tab:reward_qwen3b}, the F1-based model with action penalties not only recovers from instability, but also surpasses the EM baseline in final performance. This suggests that by ensuring active participation through direct supervision,  we can improve the stability of F1-based training and revitalize its utility for optimizing Deep Research agents.
\section{Policy Optimization}
\label{sec:algorithm}

\begin{table*}[t]
\centering
\caption{Performance comparison of different policy optimization methods based on Qwen2.5-7B.}
\label{tab:algo}
\small
\setlength{\tabcolsep}{4pt} 
\renewcommand{\arraystretch}{1.1}

\begin{tabular}{l cc @{\hspace{12pt}} cc @{\hspace{12pt}} cc @{\hspace{12pt}} cc @{\hspace{12pt}} cc}
\hline
\multirow{2}{*}{\textbf{Method}} 
& \multicolumn{10}{c}{\rule{0pt}{12pt}\textbf{Multi-Hop QA}} \\
\cmidrule(lr){2-11}
& \multicolumn{2}{c}{\textbf{HotpotQA}} & \multicolumn{2}{c}{\textbf{2Wiki}} & \multicolumn{2}{c}{\textbf{Musique}} & \multicolumn{2}{c}{\textbf{Bamboogle}} & \multicolumn{2}{c}{\textbf{Avg.}} \\
\cmidrule(lr){2-3} \cmidrule(lr){4-5} \cmidrule(lr){6-7} \cmidrule(lr){8-9} \cmidrule(lr){10-11}
& Acc & Count & Acc & Count & Acc & Count & Acc & Count & Acc & Count \\
\hline
\rule{0pt}{10pt}REINFORCE~\citep{williams1992simple}
& 0.407 & \textbf{1.44} & 0.393 & \textbf{1.56} & \textbf{0.192} & 2.00 & 0.422 & \textbf{1.70} & 0.354 & \textbf{1.68} \\
PPO~\citep{schulman2017proximal}
& \textbf{0.427} & 1.99 & 0.357 & 2.02 & 0.156 & \textbf{1.90} & \textbf{0.453} & 2.00 & 0.348 & 1.98 \\
GRPO~\citep{shao2024deepseekmath}
& 0.419 & 1.57 & \textbf{0.401} & 1.97 & 0.178 & 1.93 & 0.422 & 1.90 & \textbf{0.355} & 1.84 \\
\hline
\end{tabular}
\vspace{0.1cm} 
\begin{tabular}{l cc @{\hspace{12pt}} cc @{\hspace{12pt}} cc @{\hspace{12pt}} cc @{\hspace{12pt}} cc}
\hline
\multirow{2}{*}{\textbf{Method}} 
& \multicolumn{8}{c}{\rule{0pt}{12pt}\textbf{Single-Hop QA}} 
& \multicolumn{2}{c}{\textbf{Overall}} \\
\cmidrule(lr){2-9} \cmidrule(lr){10-11}
& \multicolumn{2}{c}{\textbf{NQ}} & \multicolumn{2}{c}{\textbf{TriviaQA}} & \multicolumn{2}{c}{\textbf{PopQA}} & \multicolumn{2}{c}{\textbf{Avg.}} & \multicolumn{2}{c}{\textbf{Avg.}} \\
\cmidrule(lr){2-3} \cmidrule(lr){4-5} \cmidrule(lr){6-7} \cmidrule(lr){8-9} \cmidrule(lr){10-11}
& Acc & Count & Acc & Count & Acc & Count & Acc & Count & Acc & Count \\
\hline
\rule{0pt}{10pt}REINFORCE~\citep{williams1992simple}
& \textbf{0.474} & \textbf{1.01} & \textbf{0.647} & 1.03 & 0.439 & \textbf{1.02} & \textbf{0.520} & \textbf{1.02} 
& \textbf{0.437} & \textbf{1.35} \\
PPO~\citep{schulman2017proximal}
& 0.463 & 1.95 & 0.641 & 1.96 & \textbf{0.455} & 1.96 & \textbf{0.520} & 1.96 
& 0.422 & 1.97 \\
GRPO~\citep{shao2024deepseekmath}
& 0.460 & \textbf{1.01} & 0.636 & \textbf{1.02} & 0.440 & 1.06 & 0.512 & 1.03 
& 0.433 & 1.44 \\
\hline
\end{tabular}
\end{table*}

\begin{wrapfigure}{r}{0.48\textwidth}
    \centering
    \includegraphics[width=0.83\linewidth]{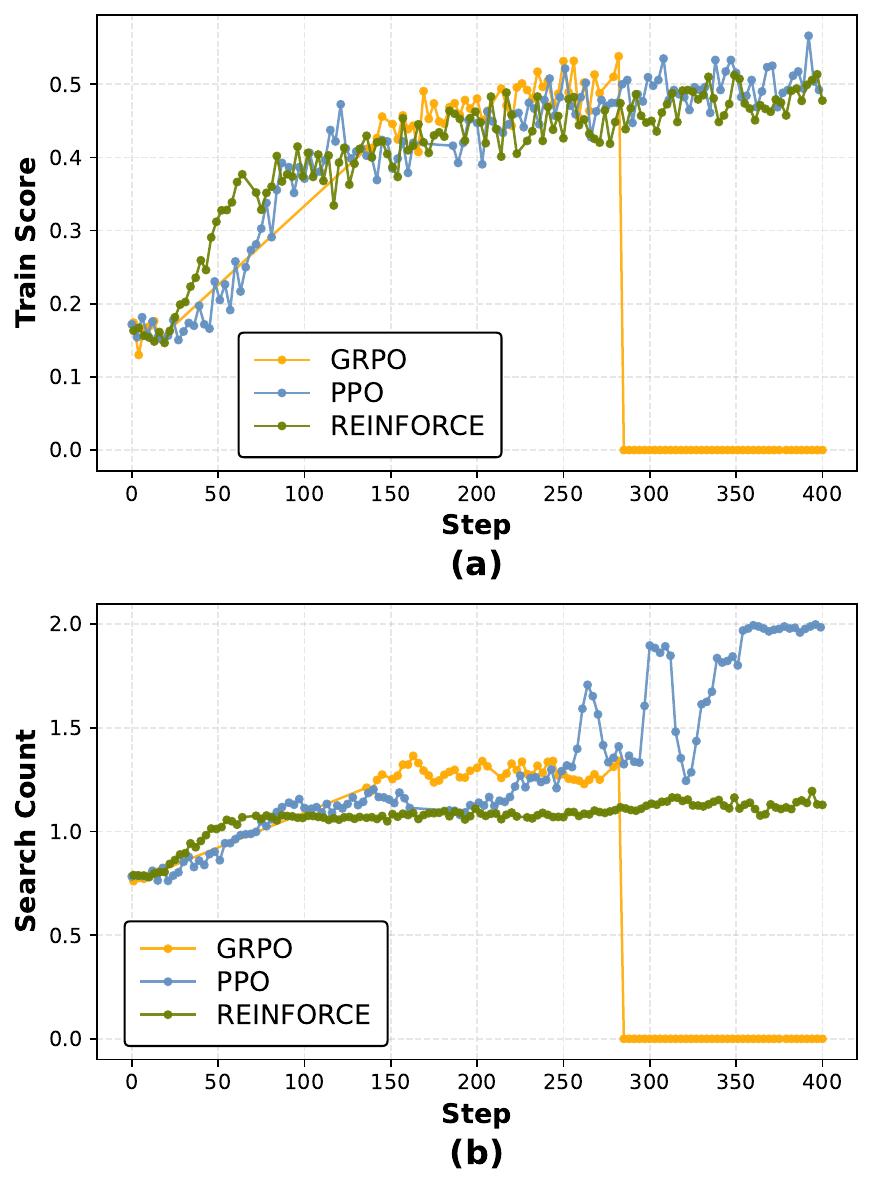}
    \vspace{-5pt}
    \caption{Training dynamics under different RL algorithms. (\textbf{a}) Training score during training. (\textbf{b}) Average number of search actions per step.}
    \label{fig:algo}
    \vspace{-15pt}
\end{wrapfigure}

\subsection{Setup}
In existing Deep Research training pipelines, PPO, GRPO, and REINFORCE are widely adopted. However, their specific algorithmic effects are often confounded by variations in prompt design, reward formulation, and engineering details, making fair comparison difficult. To isolate the policy update's role, we evaluate Qwen2.5-3B and Qwen2.5-7B using the Fast Thinking template and EM reward, with full experimental settings in Appendix~\ref{sec:detail setup}. By fixing the interaction interface and all other training parameters, we evaluate the distinct impacts of these optimization methods on stability, performance, and reasoning cost. We report results on Qwen2.5-7B in the main text, with Qwen2.5-3B results provided in Appendix~\ref{sec:appendix_3b}.

\subsection{REINFORCE vs PPO vs GRPO}
In terms of stability, as illustrated in Figure~\ref{fig:algo}, GRPO demonstrates inferior robustness in this setting, frequently suffering from training collapse, whereas both REINFORCE and PPO achieve stable convergence. 
Regarding accuracy, REINFORCE achieves the highest overall performance as shown in Table~\ref{tab:algo}. A distinct trade-off is observed between the other two: PPO outperforms GRPO on Single-Hop tasks, while GRPO exhibits superior capability on Multi-Hop benchmarks. 
Finally, concerning reasoning cost, the algorithms differ significantly in adaptability on models with stronger base capabilities. REINFORCE learns the most compact strategy with the lowest search frequency. Conversely, PPO maintains a high and rigid search count regardless of task difficulty, exhibiting similar search counts across Single and Multi-Hop tasks. This indicates a failure to adaptively reduce computational effort for simpler queries. However, we note that on the smaller Qwen2.5-3B model, due to its limited exploration capacity, all algorithms consistently perform only one single search regardless of task complexity, as detailed in Appendix~\ref{sec:appendix_3b}.

\begin{table*}[t!]
\centering
\caption{Performance comparison of different methods on single-hop and multi-hop QA benchmarks based on Qwen2.5-7B.}
\label{tab:main_results}
\small
\setlength{\tabcolsep}{5pt}
\renewcommand{\arraystretch}{1.1} 

\begin{tabular}{lcccccccc}
\hline
\multicolumn{1}{c}{\multirow{2}{*}[-0.8ex]{\begin{tabular}{@{}c@{}}\textbf{Method}\\[-2pt] \textbf{(Qwen2.5-7B)}\end{tabular}}}
& \multicolumn{3}{c}{\textbf{Single-Hop QA}} 
& \multicolumn{4}{c}{\textbf{Multi-Hop QA}} 
& \multirow{2}{*}{\textbf{Avg.}} \\
\cmidrule(lr){2-4} \cmidrule(lr){5-8}
& \textbf{NQ} & \textbf{TriviaQA} & \textbf{PopQA} 
& \textbf{HotpotQA} & \textbf{2Wiki} & \textbf{Musique} & \textbf{Bamboogle} & \\
\hline
\rowcolor{gray!15}\multicolumn{9}{c}{\rule{0pt}{10pt}RL Trained LLM without Retrieval} \\
R1-base 
& 0.297 & 0.539 & 0.202 & 0.242 & 0.273 & 0.083 & 0.296 & 0.276 \\
\hline
\rowcolor{gray!15}\multicolumn{9}{c}{\rule{0pt}{10pt}Training free Deep Research Agent} \\
ReAct 
& 0.178 & 0.276 & 0.183 & 0.132 & 0.132 & 0.039 & 0.266 & 0.172 \\
\hline
\rowcolor{gray!15}\multicolumn{9}{c}{\rule{0pt}{10pt}RL trained Deep Research Agents} \\
Search-R1~\cite{jin2025search}
& 0.451 & 0.620 & 0.434 & 0.361 & 0.386 & 0.163 & 0.406 & 0.403 \\
Search-R1++
& \textbf{0.499} & \textbf{0.672} & \textbf{0.440} & \textbf{0.423} & \textbf{0.408} & \textbf{0.205} & \textbf{0.448} & \textbf{0.442} \\
\hline
\end{tabular}
\end{table*}

\begin{table*}[t]
\centering
\caption{Performance comparison of different methods on single-hop and multi-hop QA benchmarks based on Qwen2.5-3B.}
\label{tab:main_results_3b}
\small
\setlength{\tabcolsep}{4.5pt}
\renewcommand{\arraystretch}{1.1}

\begin{tabular}{lcccccccc}
\hline
\multicolumn{1}{c}{\multirow{2}{*}[-0.8ex]{\begin{tabular}{@{}c@{}}\textbf{Method}\\[-2pt] \textbf{(Qwen2.5-3B)}\end{tabular}}}
& \multicolumn{3}{c}{\rule{0pt}{12pt}\textbf{Single-Hop QA}}
& \multicolumn{4}{c}{\textbf{Multi-Hop QA}}
& \multirow{2}{*}{\textbf{Avg.}} \\
\cmidrule(lr){2-4} \cmidrule(lr){5-8}
& \textbf{NQ} & \textbf{TriviaQA} & \textbf{PopQA}
& \textbf{HotpotQA} & \textbf{2Wiki} & \textbf{Musique} & \textbf{Bamboogle} & \\
\hline

\rowcolor{gray!15}\multicolumn{9}{c}{\rule{0pt}{10pt}RL Trained LLM without Retrieval} \\
R1-base 
& 0.226 & 0.455 & 0.173 & 0.201 & 0.268 & 0.055 & 0.224 & 0.229 \\
\hline

\rowcolor{gray!15}\multicolumn{9}{c}{\rule{0pt}{10pt}Training free Deep Research Agent} \\
ReAct 
& 0.063 & 0.130 & 0.072 & 0.047 & 0.046 & 0.008 & 0.016 & 0.055 \\
\hline

\rowcolor{gray!15}\multicolumn{9}{c}{\rule{0pt}{10pt}RL trained Deep Research Agents} \\
Search-R1~\cite{jin2025search}
& 0.396 & 0.570 & 0.381 & 0.263 & 0.254 & 0.048 & 0.109 & 0.289 \\
Search-R1++
& \textbf{0.427} & \textbf{0.608} & \textbf{0.432} & \textbf{0.325} & \textbf{0.300} & \textbf{0.065} & \textbf{0.162} & \textbf{0.331} \\
\hline
\end{tabular}
\end{table*}

These behavioral differences stem from how each algorithm incorporates baselines and handles credit assignment. To elaborate, both PPO and GRPO introduce auxiliary mechanisms to reduce variance, with PPO utilizing a value model and GRPO employing group averaging. However, these mechanisms introduce instability or bias in Deep Research.
\textbf{GRPO} relies on relative advantages within sampled groups. In multi-step, long-context reasoning, the high variance of actions within a group makes the baseline noisy, leading to training instability.
\textbf{PPO} depends on a learned critic for advantage estimation. In scenarios with sparse outcome rewards (EM), fitting an accurate value function over long trajectories is challenging. This difficulty often leads to critic bias that fails to penalize redundant searches, explaining PPO's high and unadaptive search cost.
\textbf{REINFORCE}, by contrast, optimizes the policy based on the direct cumulative return without relying on external baseline. By avoiding the noise from group sampling and the bias from critic estimation, it learns the most efficient search-and-answer path, resulting in the best stability and lowest reasoning cost.

\section{A strong baseline: Search-R1++}

We distill the core insights from our experimental analysis into the following three conclusions:
\begin{itemize}
    \setlength{\itemsep}{0pt} 
    \setlength{\parskip}{1pt} 
    \item \textbf{Prompt Template:} The longer the explicit reasoning length, the worse the accuracy and training stability. So we adopt the Fast Thinking template to ensure robust convergence and higher accuracy.

    \item \textbf{Reward Fuction:} F1-based training performs worse than EM and exhibits instability caused by the policy degenerating into answer avoidance. To address this, we introduce F1+, which augments F1 with penalties for no search or answers, to stabilize training and outperform the EM-based approach.
    
    \item \textbf{Policy Optimization:} GRPO suffers from instability, whereas REINFORCE achieves superior stability and efficiency by avoiding baseline interference.
\end{itemize}


Based on the above insights, we develop \textbf{Search-R1++} using the Fast Thinking template and REINFORCE with F1+ rewards. To evaluate its effectiveness, we compare it against the following baselines:
(1) RL Trained LLM without Retrieval: R1-base is trained via PPO on the same data, performing reasoning without search engine. 
(2) Training Free Deepresearch Agent: ReAct, a training-free agent using the identical inference pipeline as ours for direct inference. (3) Deep Research Agents trained with RL: 
Search-R1~\citep{jin2025search}. For Search-R1, despite our efforts to reproduce the experiments, we cannot fully match the reported performance. This observation aligns with findings from other independent reproductions~\citep{sun2025zerosearch}.

As reported in Tables~\ref{tab:main_results} and~\ref{tab:main_results_3b}, we achieve 3.9\% and 4.2\% average relative improvement with Qwen2.5-7B and Qwen2.5-3B respectively, compared to the Search-R1 baseline. While training-free methods degrade significantly on smaller models, our \textbf{Search-R1++} approach maintains robust performance across scales, demonstrating that a principled RL strategy effectively empowers even compact architectures.

\section{Conclusion}
This work presents the first comprehensive study of reinforcement learning for Deep Research agents from a unified perspective. 
By systematically analyzing three decoupled dimensions: \textbf{prompt template}, \textbf{reward function}, and \textbf{policy optimization}, we identify critical training settings impacting agent performance across answer accuracy, training stability, and inference cost.
Building on these insights, we propose ~\textbf{Search-R1++}, a strong baseline adopting the Fast Thinking template and trained via REINFORCE with F1+ reward. 
As Deep Research represents a prototypical setting for long-horizon LLM reasoning, our findings also provide guidance for developing more principled RL training strategies for Large Language Models broadly.

\bibliography{main}

\newpage 
\appendix 
\onecolumn
\appendix
\section{Appendix}

\begin{figure*}[t]
    \centering
    \includegraphics[width=0.95\textwidth]{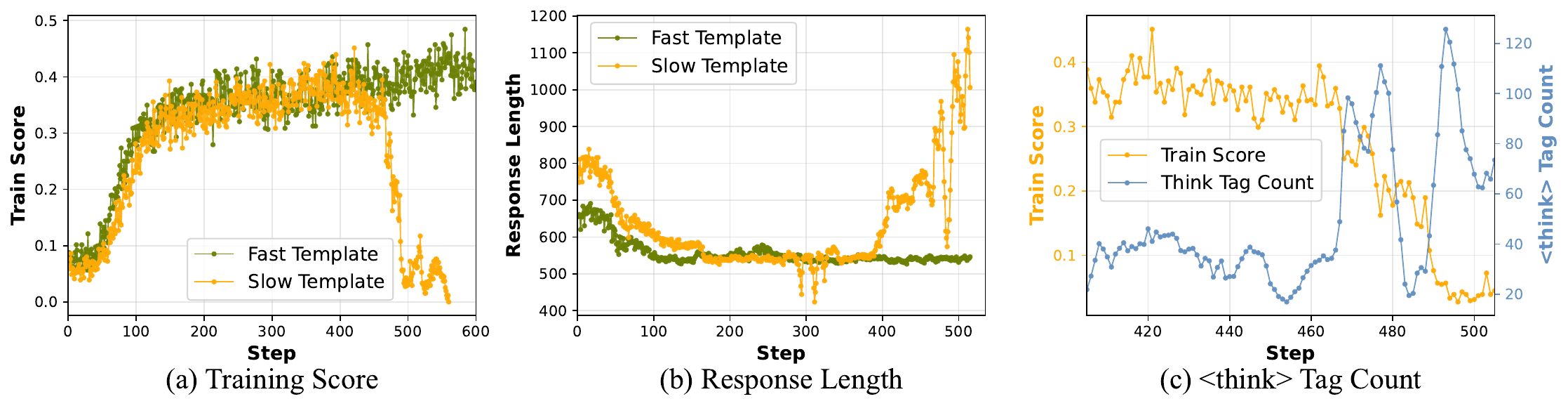}
    \caption{Training dynamics of Qwen2.5-3B under different prompt templates.(\textbf{a}) compares the training score under Fast and Slow Thinking templates; (\textbf{b}) shows the average response length evolution over training steps; (\textbf{c}) illustrates the surge in \texttt{<think>} tags coinciding with the performance collapse.}
    \label{fig:prompt-2-3b}
\end{figure*}

\begin{figure}[t]
    \centering
    \begin{minipage}[t]{0.48\linewidth}
        \centering
        \includegraphics[width=\linewidth]{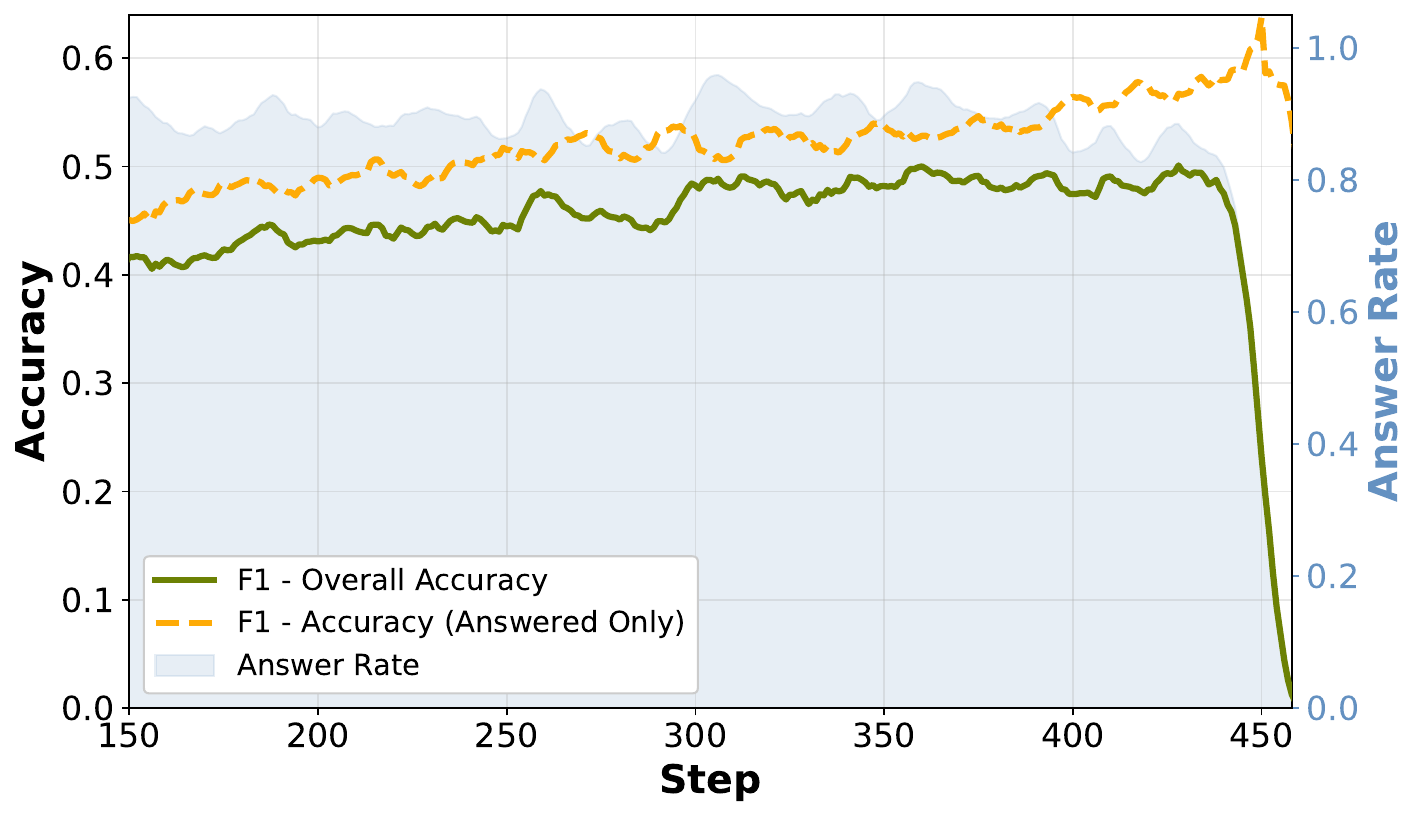}
        \caption{Overall accuracy, answered-only accuracy, and answer rate (shaded area) under F1 reward training for Qwen2.5-3B.}
        \label{fig:reward-1-3b}
    \end{minipage}
    \hfill
    \begin{minipage}[t]{0.48\linewidth}
        \centering
        \includegraphics[width=\linewidth]{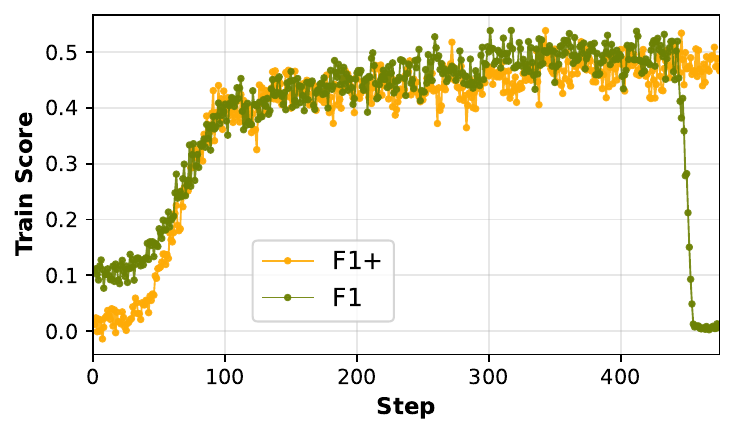}
        \caption{Training score under F1 and F1+ reward for Qwen2.5-3B. F1+ denotes F1 reward augmented with action supervision.}
        \label{fig:reward-2-3b}
    \end{minipage}
\end{figure}

\subsection{Complementary Experimental Results on Qwen2.5-3B}
\label{sec:appendix_3b}

In this section, we provide additional experimental results based on Qwen2.5-3B to demonstrate the generalizability of our findings. Unless otherwise noted, the experimental setup remains identical to that described in the main text.

\textbf{Prompt Templates.} Consistent with the observations on Qwen2.5-7B in Section~\ref{sec:prompt}, the Qwen2.5-3B model exhibits nearly identical training dynamics. As shown in Figure~\ref{fig:prompt-2-3b}, the Fast Thinking template maintains robust stability throughout training, whereas the Slow Thinking template is prone to collapse. This instability is similarly characterized by a sudden surge in both response length and the frequency of \texttt{<think>} tags blocks that hinder effective decision-making.

\textbf{Reward Function.} Consistent with the observations on Qwen2.5-7B in Section~\ref{sec:reward}, the Qwen2.5-3B model exhibits nearly identical training dynamics under different reward formulations. As shown in Figure~\ref{fig:reward-1-3b}, the F1-based training demonstrates the same answer avoidance pattern: sharp drops in overall accuracy coincide with significant declines in answer rate, while the accuracy of answered samples remains relatively stable. This confirms that the dominant failure mode is answer refusal rather than incorrect answering. Furthermore, as shown in Figure~\ref{fig:reward-2-3b}, F1+ successfully stabilizes the training process by eliminating this refusal behavior, mirroring the effectiveness observed in the Qwen2.5-7B model.

\begin{table*}[t]
\centering
\caption{Performance comparison of different optimization algorithms based on Qwen2.5-3B.}
\label{tab:algo_qwen3b}
\vspace{0.1in}
\small
\setlength{\tabcolsep}{4.5pt}
\renewcommand{\arraystretch}{1.1}
\begin{tabular}{lcccccccc}
\hline
\multirow{2}{*}{\textbf{Method}} 
& \multicolumn{3}{c}{\rule{0pt}{10pt}\textbf{Single-Hop QA}}
& \multicolumn{4}{c}{\textbf{Multi-Hop QA}}
& \multirow{2}{*}{\textbf{Avg.}} \\
\cmidrule(lr){2-4} \cmidrule(lr){5-8}
& \textbf{NQ} & \textbf{TriviaQA} & \textbf{PopQA}
& \textbf{HotpotQA} & \textbf{2Wiki} & \textbf{Musique} & \textbf{Bamboogle} & \\
\hline
\rule{0pt}{12pt}REINFORCE
& \textbf{0.438} & \textbf{0.604} & \textbf{0.447} & \textbf{0.317} & 0.284 & \textbf{0.066} & \textbf{0.141} & \textbf{0.328} \\
PPO
& 0.390 & 0.576 & 0.393 & 0.282 & 0.272 & 0.041 & 0.125 & 0.297 \\
GRPO
& 0.415 & 0.586 & 0.439 & 0.306 & \textbf{0.297} & 0.062 & 0.135 & 0.315 \\
\hline
\end{tabular}
\end{table*}

\begin{wrapfigure}[15]{R}{0.48\textwidth} 
    \centering
    \vspace{-10pt} 
    \includegraphics[width=\linewidth]{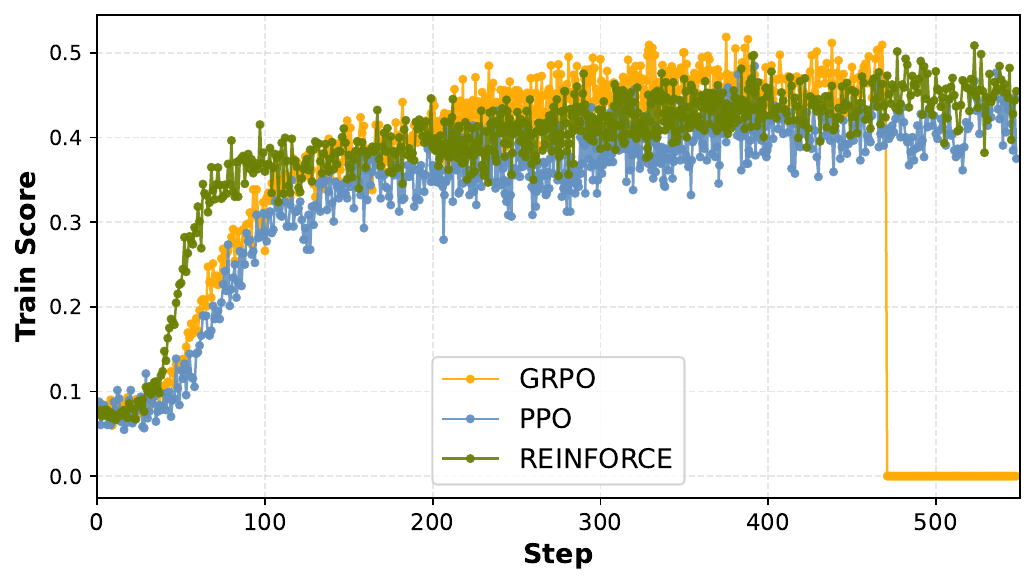}
    \caption{Training score of Qwen2.5-3B under different RL algorithms.}
    \label{fig:algo_3b}
\end{wrapfigure}
\textbf{Policy Optimization}. Consistent with the observations on Qwen2.5-7B in Section~\ref{sec:algorithm}, the Qwen2.5-3B model exhibits similar stability patterns under different optimization algorithms. As shown in Figure~\ref{fig:algo_3b}, GRPO demonstrates inferior robustness and is prone to training collapse, whereas both REINFORCE and PPO achieve stable convergence. Regarding performance, REINFORCE consistently achieves the highest overall accuracy, mirroring the trend observed in the larger model. A notable divergence, however, lies in the reasoning cost. Unlike the 7B model, Qwen2.5-3B converges to a rigid strategy of executing only a single search step across all algorithms and datasets. Consequently, we do not report the search count in Table~\ref{tab:algo_qwen3b}. This behavior likely stems from the limited instruction-following and long-horizon reasoning capabilities of the smaller base model, which struggles to effectively leverage multi-turn interactions or explore complex search trajectories, instead defaulting to the most basic retrieval pattern.

\subsection{Case Study}
\label{sec:case}

In Section~\ref{sec:prompt} of the main text, we analyzed how prompt design critically influences the optimization landscape and training stability of Deep Research agents. We observed that the explicit Slow Thinking Template often introduces unfavorable optimization biases, leading to progressive policy degeneration, whereas our proposed Fast Thinking Template maintains robust training dynamics. To provide a concrete understanding of these behaviors, we present four representative examples generated by Qwen-2.5-7B, illustrating the trajectories under different settings:

\begin{table*}
\centering
\vspace{0.1in}
\caption{An example of Slow Thinking Template during normal reasoning.}
\label{tab:slow-1}
\vspace{0.1in}
\begin{tabular}{p{0.95\textwidth}}
\toprule
\textbf{Question:} Who has defined function of management as POSDCORB? \\
\midrule
\textbf{Ground Truth:} Luther Gulick \\
\midrule
\textbf{Slow Thinking Template:} \\
\textcolor{violet}{\textless think\textgreater} The definition of the function of management by POSDCORB is in Wikipedia. So I conduct the following search \textcolor{violet}{\textless /think\textgreater} \\
\textcolor{cyan}{\textless search\textgreater} POSDCORB function of management \textcolor{cyan}{\textless /search\textgreater} \\
\textcolor{orange}{\textless information\textgreater} Doc 1(Title: POSDCORB) Committee, Luther Gulick asks rhetorically "What is the work of the chief executive? What does he do?" POSDCORB is the answer, "designed to call attention to the various functional elements of the work of a chief executive because 'administration' and 'management' have lost all specific content." According to Gulick, the elements are: Gulick's "Notes on the Theory of Organization" further ... \textcolor{orange}{\textless /information\textgreater} \\
\textcolor{violet}{\textless think\textgreater} As per information 1, the description and definition of POSDCORB are found in Luther Gulick's notes on the theory of organization. The main elements included in this acronym are Planning, Organizing, Staffing, Directing, Co-ordinating, Reporting and Budgeting. These tasks, Gulick states, together form the primary functions of a chief executive. \textcolor{violet}{\textless /think\textgreater} \\
\textcolor{red}{ \textless answer\textgreater } Luther Gulick \textcolor{red}{\textless /answer\textgreater} \\
\bottomrule
\end{tabular}
\end{table*}

\textbf{Example 1: Slow Thinking (Early Stage / Normal Behavior).}
Table~\ref{tab:slow-1} displays a generation sample from the early training stage of the Slow Thinking model. At this point, the agent produces coherent reasoning steps (wrapped in \texttt{<think>} tags) followed by a correct search-and-answer process, aligning with the expected behavior.

\newpage
\begin{table*}
\centering
\vspace{0.1in}
\caption{An example of Slow Thinking Template before the collapse phase with increasing meaningless thinking.}
\label{tab:slow-2}
\vspace{0.1in}
\begin{tabular}{p{0.95\textwidth}}
\toprule
\textbf{Question:} When's the last time the Jaguars made the playoffs? \\
\midrule
\textbf{Ground Truth:} 2017 \\
\midrule
\textbf{Slow Thinking Template:} \\
\textcolor{violet}{\textless think\textgreater} Jacksonville Jaguars \textcolor{violet}{\textless /think\textgreater} \\
\textcolor{violet}{\textless think\textgreater} last playoff appearance \textcolor{violet}{\textless /think\textgreater} \\
\textcolor{violet}{\textless think\textgreater} 2007 \textcolor{violet}{\textless /think\textgreater} \\
\textcolor{violet}{\textless think\textgreater} To answer the question, I will search for the last time the Jacksonville Jaguars made the playoffs. \textcolor{violet}{\textless /think\textgreater} \\
\textcolor{cyan}{\textless search\textgreater} last time jacksonville jaguars made the playoffs \textcolor{cyan}{\textless /search\textgreater} \\
\textcolor{orange}{\textless information\textgreater} Doc 1(Title: "Jacksonville Jaguars") Jacksonville Jaguars The Jacksonville Jaguars are an American professional football franchise based in Jacksonville, Florida. The Jaguars compete in the National Football League (NFL) as a member club of the American Football Conference (AFC) South division. The team plays its home games at TIAA Bank Field. The Jaguars and the Carolina Panthers joined the NFL as expansion teams for the 1995 season... \textcolor{orange}{\textless /information\textgreater} \\
\textcolor{violet}{\textless think\textgreater} 2017 \textcolor{violet}{\textless /think\textgreater} \\
\textcolor{violet}{\textless think\textgreater} \textcolor{violet}{\textless /think\textgreater} \\
\textcolor{red}{\textless answer\textgreater} 2017 \textcolor{red}{\textless /answer\textgreater} \\
\bottomrule
\end{tabular}
\end{table*}

\textbf{Example 2: Slow Thinking (Intermediate Stage / Signs of Instability).} 
Table~\ref{tab:slow-2} illustrates early signs of instability. While the agent still generates correct answers, it begins to output meaningless or repetitive \texttt{<think>} tags before deciding on an action. This indicates that the reasoning process is becoming redundant, even if the final outcome remains correct.

\begin{table*}
\centering
\vspace{0.1in}
\caption{An example of Slow Thinking Template after collapse with repetitive thinking loops.}
\label{tab:slow-3}
\vspace{0.1in}
\begin{tabular}{p{0.95\textwidth}}
\toprule
\textbf{Question:} When did the first Curious George book come out? \\
\midrule
\textbf{Ground Truth:} 1939 \\
\midrule
\textbf{Slow Thinking Template:} \\
\textcolor{cyan}{\textless search\textgreater} when did the first curious george book come out? \textcolor{cyan}{\textless /search\textgreater} \\
\textcolor{orange}{\textless information\textgreater} Doc 1(Title: "Curious George (book)") Curious George (book) Curious George is a children's book written and illustrated by Margret Rey and H. A. Rey, and published by Houghton Mifflin in 1941. It is the first book in the Curious George series... \textcolor{orange}{\textless /information\textgreater} \\
\textcolor{violet}{\textless think\textgreater} 1941 \textcolor{violet}{\textless /think\textgreater} \\
\textcolor{violet}{\textless think\textgreater} \textless think\textgreater 1941 \textless /think\textgreater \textcolor{violet}{\textless /think\textgreater} \\
\textcolor{violet}{\textless think\textgreater} 1941 \textcolor{violet}{\textless /think\textgreater} \\
\textcolor{violet}{\textless think\textgreater} Curious George \textcolor{violet}{\textless /think\textgreater} \\
\textcolor{violet}{\textless think\textgreater} 1941 \textcolor{violet}{\textless /think\textgreater} \\
\textcolor{violet}{\textless think\textgreater} \textless think\textgreater 1941 \textless /think\textgreater \textcolor{violet}{\textless /think\textgreater} \\
\textcolor{violet}{\textless think\textgreater} \textless think\textgreater 1941 \textless /think\textgreater \textcolor{violet}{\textless /think\textgreater} \\
\textcolor{violet}{\textless think\textgreater} \textless think\textgreater 1941 \textless /think\textgreater \textcolor{violet}{\textless /think\textgreater} \\
... \\
My previous action is invalid. If I want to search, I should put the query between \textless search\textgreater and \textless /search\textgreater. If I want to give the final answer, I should put the answer between \textless answer\textgreater and \textless /answer\textgreater. Let me try again. \\
\textcolor{violet}{\textless think\textgreater} 1941 \textcolor{violet}{\textless /think\textgreater} \\
\textcolor{violet}{\textless think\textgreater} 1941 \textcolor{violet}{\textless /think\textgreater} \\
... \\
\bottomrule
\end{tabular}
\end{table*}

\textbf{Example 3: Slow Thinking (Late Stage / Mode Collapse).} 
Table~\ref{tab:slow-3} demonstrates the final stage of training collapse under the Slow Thinking Template. The agent fails to execute valid search actions or generates refusal responses (e.g., \textit{"I cannot answer..."}) despite having the capability, indicating that the policy has fallen into a degenerate local optimum of answer avoidance.

\begin{table*}[!htb]
\centering
\vspace{0.1in}
\caption{An example of Fast Thinking Template }
\label{tab:fast}
\vspace{0.1in}
\begin{tabular}{p{0.95\textwidth}}
\toprule
\textbf{Question:} When did the capital of Nigeria moved from Lagos to Abuja? \\
\midrule
\textbf{Ground Truth:} 12 December 1991 \\
\midrule
\textbf{Fast Thinking Template:} \\
\textcolor{cyan}{\textless search\textgreater} when did the capital of nigeria moved from lagos to abuja \textcolor{cyan}{\textless/search\textgreater} \\
\textcolor{orange}{\textless information\textgreater} Doc 1(Title: Abuja) Abuja Abuja () is the capital city of Nigeria located in the centre of the country within the Federal Capital Territory (FCT). It is a planned city and was built mainly in the 1980s, replacing the country's most populous city of Lagos as the capital on 12 December 1991. Abuja's geography is defined by Aso Rock, a monolith left by water erosion. The Presidential Complex, National Assembly, Supreme Court and much of the city extend to the south of the rock. Zuma Rock... \\
\textcolor{red}{\textless answer\textgreater} 12 December 1991 \textcolor{red}{\textless/answer\textgreater} \\
\bottomrule
\end{tabular}
\end{table*}

\textbf{Example 4: Fast Thinking (Stable Behavior).} 
In contrast, Table~\ref{tab:fast} presents a sample from the model trained with the Fast Thinking Template. The agent bypasses the verbose reasoning simulation and directly outputs concise, decision-oriented search queries and answers. This approach yields a stable policy that focuses on effective information retrieval without structural degeneration.

\subsection{Detailed Experiment Setup}
\label{sec:detail setup}
To facilitate reproducibility and strictly evaluate the algorithmic effectiveness, our experimental configurations are aligned with established baselines. Specifically, the hyperparameter settings for PPO and GRPO follow the protocols of SEARCH-R1~\citep{jin2025search}, while the REINFORCE implementation adopts settings consistent with ZeroSearch~\citep{sun2025zerosearch}. To ensure a fair comparison, we maintain identical computational environments and hyperparameters across all three methods wherever applicable.

To elaborate, the retrieval system relies on the E5 model~\citep{wang2022text} to query the 2018 Wikipedia snapshot~\citep{karpukhin2020dense}, retrieving the top-3 relevant passages for all experiments. For all training experiments, we merge the training sets of NQ and HotpotQA as the unified training dataset. Evaluation is conducted on the test or validation sets of seven datasets to assess both in-domain and out-of-domain performance.

Across all three algorithms, training is performed on a single node equipped with 8 A100 GPUs for a total of 600 steps. We employ a global batch size of 512, with a mini-batch size of 256 and a micro-batch size of 64. The policy model is initialized with a learning rate of 1e-6 and a consistent warm-up ratio of 0.285. The context window is configured with a maximum sequence length of 4,096 tokens, allocating up to 500 tokens for responses and 500 tokens for the top-3 retrieved passages. To optimize memory efficiency, we utilize the AdamW optimizer with Fully Sharded Data Parallel (FSDP), enabling CPU offloading for parameters and optimizer states, alongside gradient checkpointing. Efficient rollouts are generated using vLLM with a tensor parallel size of 1, a GPU memory utilization of 0.6, and a sampling temperature of 1.0. Model checkpoints are saved every 100 steps.

Regarding algorithm-specific configurations, PPO incorporates a value network (critic) initialized with a learning rate of 1e-5 and a warm-up ratio of 0.015. We utilize Generalized Advantage Estimation (GAE) for this method and sample a single response per prompt. For the GRPO and REINFORCE baselines, we eliminate the critic and sample 5 responses per prompt (n\_agent=5) to estimate the baseline. Both of these methods utilize a KL divergence coefficient of 0.001.

\subsection{Details of Pearson Correlation Calculation}
\label{sec:person}

To quantitatively analyze the relationship between explicit reasoning behavior and reward signals during training, we perform a sample-level Pearson correlation analysis. This analysis aims to reveal whether the reward mechanism inadvertently encourages the model to generate redundant reasoning markers.

\noindent\textbf{Variable Extraction and Pre-processing.}
We extract two core metrics from the model rollouts for each sample:
\begin{itemize}
    \item \textbf{Reasoning Complexity ($N_{\text{think}}$)}: The total count of \texttt{<think>} tags generated within a single response.
    \item \textbf{Immediate Reward ($R$)}: The raw score assigned to the sample by the reward model or environment.
\end{itemize}
Prior to calculation, we perform data cleaning to remove non-standard samples containing missing values (NaN) to ensure statistical validity.

\noindent\textbf{Window Selection and Data Aggregation.}
To capture the precursors of training collapse, we employ an aggregation strategy within a specific temporal window. We define the \textit{collapse start step} and trace back $k$ steps (where $k=100$ in our experiments). All samples $(N_{\text{think}, i}, R_i)$ generated within this interval $[t_{\text{collapse}}-k, t_{\text{collapse}})$ are aggregated to form high-dimensional vectors, allowing us to evaluate the correlation at a statistically significant scale.

\noindent\textbf{Quantile Binning and Smoothing.}
To mitigate noise from individual sample variance and observe macro-trends, we apply Quantile Binning (equal-frequency binning). We divide the distribution of $N_{\text{think}}$ into $m$ discrete bins based on sample frequency. For each bin, we calculate the mean reward $\bar{R}_{\text{bin}}$. This smoothing technique helps identify whether increases in reasoning segments consistently align with higher reward expectations despite the inherent stochasticity of RL training.

\noindent\textbf{Correlation Formula.}
The Pearson correlation coefficient $\rho$ is calculated on the raw sample-level data as follows:
\begin{equation}
\rho_{N, R} = \frac{\sum_{i=1}^{n} (N_i - \bar{N})(R_i - \bar{R})}{\sqrt{\sum_{i=1}^{n} (N_i - \bar{N})^2 \sum_{i=1}^{n} (R_i - \bar{R})^2}}
\end{equation}
where $n$ is the total number of samples in the aggregation window, and $\bar{N}, \bar{R}$ represent the sample means of the reasoning tag counts and rewards, respectively.

\end{document}